\documentclass{article}
\usepackage{spconf,amsmath,graphicx}
\usepackage{textcomp}
\usepackage{xcolor}

\def\L{{\cal L}}

\title{Salient object detection on hyperspectral images \\using features learned from unsupervised segmentation task}
\name{N. \.{I}mamo\u{g}lu\textsuperscript{1,*},
         G. Ding\textsuperscript{1,2},          
         Y. Fang\textsuperscript{2}, 
         A. Kanezaki\textsuperscript{1},
         T. Kouyama\textsuperscript{1}, 
         R. Nakamura\textsuperscript{1}
         \thanks{ \textsuperscript{*}Corresponding Author: Nevrez \.{I}mamo\u{g}lu (e-mail: nevrez.imamoglu@aist.go.jp). Guanqun Ding and Nevrez \.{I}mamo\u{g}lu have contributed to this work equally. This paper is based on the results obtained from a project commissioned by the New Energy and Industrial Technology Development Organization (NEDO), Japan }
}

\address{\textsuperscript{1}National Institute of Advanced Industrial Science and Technology (AIST), \\
             Artificial Intelligence Research Center, Tokyo JAPAN\\
            \textsuperscript{2}Jiangxi University of Finance and Economics, \\
            School of Information Technology, Nanchang CHINA\\
            \\
            \emph{To appear in IEEE ICASSP 2019 (accepted version)}}

\begin{document}
%
\maketitle
\begin{abstract}
Various saliency detection algorithms from color images have been proposed to mimic eye fixation or attentive object detection response of human observers for the same scenes. However, developments on hyperspectral imaging systems enable us to obtain redundant spectral information of the observed scenes from the reflected light source from objects. A few studies using low-level features on hyperspectral images demonstrated that salient object detection can be achieved. In this work, we proposed a salient object detection model on hyperspectral images by applying manifold ranking (MR) on self-supervised Convolutional Neural Network (CNN) features (high-level features) from unsupervised image segmentation task. Self-supervision of CNN continues until clustering loss or saliency maps converges to a defined error between each iteration. Finally, saliency estimations is done as the saliency map at last iteration when the self-supervision procedure terminates with convergence. Experimental evaluations demonstrated that proposed saliency detection algorithm on hyperspectral images is outperforming state-of-the-arts hyperspectral saliency models including the original MR based saliency model.
\end{abstract}
\section{Introduction}
\label{sec:intro}

Reflected light from objects and radiance from a light source has information at various wavelengths both in visible and non-visible spectrum to human eye \cite{21,6,1,2,23}. With the recent advancements on hyperspectral imaging systems, unlike the conventional cameras providing images with 1(monochrome images) or 3 channel (e.g. RGB or YCbCr images), hyperspectral imaging systems enable researchers the opportunity to capture data from the observed scenes with high spatial and redundant spectral resolution (both visible and non-visible spectrum to human eye) of the observed scenes from the radiance or reflected light source from objects  \cite{6,1,2,23}. These data have been used in many applications (remote sensing \cite{6,5,7}, scene analysis or object detection \cite{6,5,3,4,7,8,9}, spectral estimation\cite{9,10,11,12}, etc.).

Visual attention modeling (saliency detection) \cite{13,14,24,16,17} is a promising research field for practical applications, which may benefit many other applications on hyperspectral data processing stated prior. For instance, a few studies using low-level features on hyperspectral images demonstrated that salient object detection can be achieved \cite{3,4,7,8}. In contrast to these models relying on low-level features or hand-crafted features to obtain saliency maps, higher-level features can be extracted and used in a self-supervised manner for hyper-spectral data, where each spectral bands' contribution to the representation can be learned with unsupervised neural network used for segmentation task \cite{25}. In addition, works on hyperspectral saliency on natural scenes were mostly tested on dataset with a few hyperspectral images (\cite{3} used 13 images and \cite{4} used 17 images) collected and selected from various hyperspectral data. Moreover, these hyperspectral data was not collected and created for the purpose of salient object detection. And, quantitative evaluations of the models were mostly limited to Precision -Recall and F-measure metrics. Therefore, we believe that a dataset created specifically for salient object detection should be used for evaluating the models with various metrics. 

\begin{figure*}[ht]
\begin{minipage}[b]{1.0\linewidth}
  \centering
  \centerline{\includegraphics[width=15.5cm]{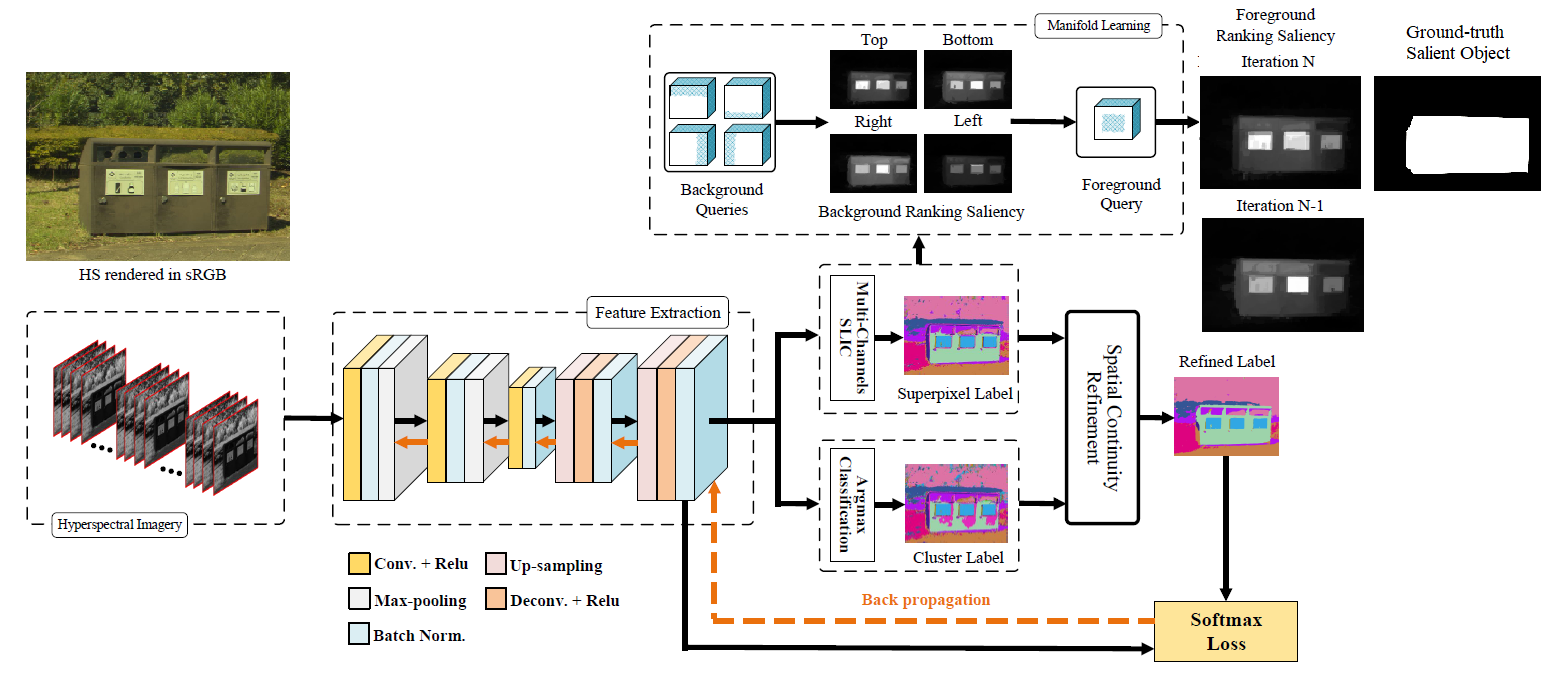}}
\end{minipage}
\caption{Proposed hyperspectral salient object detection model with unsupervised deep features .}
\label{fig:model}
\end{figure*}

\textbf{Proposed work and contributions:} In this work, we propose a salient object detection model (see Figure 1) on hyperspectral images by applying manifold ranking \cite{24} to self-supervised Convolutional Neural Network (CNN) features (high-level features) learned by an unsupervised image segmentation task \cite{25}. Self-supervision of CNN continues until clustering loss or saliency map computed from CNN features converges to a defined error between each iteration. Then, saliency map at the last iteration is used as the result of proposed model when the self-supervision procedure terminates. 

\begin{table}
\centering \caption{The detailed configuration of the CNN model. Note that \emph{BN} represents batch normalization operation.}
\begin{tabular}{  c | c | c | c  }

\hline
layer & \# filters & kernel & stride \\
\hline\hline
Conv. + Relu + BN & 64 & 3 $\times$ 3 & 1 $\times$ 1 \\
Maxpooling & - & 2 $\times$ 2 & 2 $\times$ 2 \\
\hline
Conv. + Relu + BN & 64 & 3 $\times$ 3 & 1 $\times$ 1 \\
Maxpooling & - & 2 $\times$ 2 & 2 $\times$ 2 \\
\hline
Conv. + Relu + BN & 64 & 3 $\times$ 3 & 1 $\times$ 1 \\
\hline
Upsampling & - & 2 $\times$ 2 & 2 $\times$ 2 \\
Deconv. + Relu + BN & 64 & 3 $\times$ 3 & 1 $\times$ 1 \\
\hline
Upsampling & - & 2 $\times$ 2 & 2 $\times$ 2 \\
Deconv. + Relu + BN & 64 & 1 $\times$ 1 & 1 $\times$ 1 \\
\hline

\end{tabular}
\label{tab:CNN_model}
\end{table}

To the best of our knowledge, there are not any works on hyperspectral salient object detection for natural scenes as a self-supervised approach, which combines unsupervised segmentation using CNN and salient object detection task on the scene. Regarding the approach, contributions or differences of the proposed model can be explained as: First, unsupervised segmentation task used in previous paper \cite{25} takes advantage of cluster refinement process based on the superpixels obtained by the input color image. However, in this work, we apply the refinement process based on the superpixels obtained by the high-level features of the CNN (see Fig.1) that takes hyperspectral image as input. Interestingly, this process resulted in better saliency detection performance and it seemed faster convergence regarding the segmentation task. Second, in contrast to the saliency model with manifold ranking (MR) in \cite{24} using low-level features, we utilized self-supervised deep-features with higher order semantics, which seems to improve the saliency detection performance drastically compared to study in \cite{24}. Then, unlike the CNN model used in \cite{25}, we included max-pooling for down-sampling and we replaced last two CNN layers with deconvolution layers as in Table \ref{tab:CNN_model}. Finally, self-supervision of CNN model does not need to finalize until a defined maximum iteration because we check clustering loss and saliency map for termination; in addition, saliency results of proposed model seems to converge faster than the segmentation task in most cases while using self-supervised deep features on manifold ranking based saliency detection. Experiments demonstrated that proposed saliency detection algorithm on hyperspectral images is outperforming state-of-the-arts hyperspectral saliency models including the original MR \cite{24} saliency model.

\section{Self-supervised salient object detection on hyperspectral images}
\label{sec:hs_saliency}

\begin{table*}[!ht]
\centering \caption{The performance of different saliency detection methods are given as values with Red, Green, and Blue colors indicate the best three results in respective order. Note that larger AUC$_{Borji}$, CC, F$_{\beta}$, maxF$_{\beta}$, aveF$_{\beta}$, Precision, Recall, NSS values, and smaller KLdiv values means better performance.}
\begin{tabular}{ | c | c | c | c | c | c | c | c | c | c | c }
\hline
\small Model              & \small AUC$_{Borji}$ &  \small CC   &  \small F$_{\beta}$  & \small  maxF$_{\beta}$   & \small  aveF$_{\beta}$  & \small  Precision  & \small  Recall & \small NSS & \small KLdiv   \\
\hline\hline
\small Itti \emph{et al.}  \cite{13,3} & 0.7774     & 0.3536 & 0.3530      & 0.3754 & 0.1674 & 0.1909   & 0.2329 & 1.3636 & 2.3186 \\
\small SED  \cite{3}               & 0.7691    & 0.2797 & 0.3082      & 0.3420 & 0.1541 & 0.3479   & 0.1207 & 1.3498 & 2.3426 \\
\small SAD  \cite{3}              & 0.7707    & 0.3034 & 0.2635      & 0.2662 & 0.1397 & 0.1547   & 0.2314 & 1.1767 & 2.2719 \\
\small GS   \cite{3}              & 0.7781    & 0.3403 & 0.3004      & 0.3694 & 0.1861 & 0.2753   & 0.2275 & 1.5637 & 2.1944 \\
\small SED-OCM-GS \cite{3}        & 0.8021     & 0.3730 & 0.3260      & 0.3634 & 0.1708 & 0.2757   & 0.2209 & \textcolor{blue}{\textbf{1.5908}} & 2.1707 \\
\small SED-OCM-SAD \cite{3}        & 0.8108     & 0.3882 & 0.2635      & 0.2662 & 0.1397 & 0.1547   & 0.2314 & 1.5301 & \textcolor{blue}{\textbf{2.1601}} \\
\small SGC \cite{4}*   &  \textcolor{blue}{\textbf{0.8252}}  & \textcolor{blue}{\textbf{0.5012}} & 0.1891  & 0.2214 & 0.1815 & 0.2344   & 0.2822 & 1.4739 & 2.2154 \\
\small HS-MR  \cite{24}**    & 0.7369   & 0.3492 & \textcolor{blue}{\textbf{0.3636}} & \textcolor{blue}{\textbf{0.4308}} & \textcolor{blue}{\textbf{0.3638}} & \textcolor{blue}{\textbf{0.4397}}  & \textcolor{blue}{\textbf{0.3587}} & 1.2702 & 3.7637 \\
\hline

\textbf{\small SUDF$_{HS-Slic}$}  & \textcolor{green}{\textbf{0.8509}}  & \textcolor{green}{\textbf{0.5563}} & \textcolor{green}{\textbf{0.4580}}  & \textcolor{green}{\textbf{0.5355}} & \textcolor{green}{\textbf{0.4430}} & \textcolor{green}{\textbf{0.5346}} & \textcolor{green}{\textbf{0.4449}} & \textcolor{red}{\textbf{2.1938}} & \textcolor{green}{\textbf{1.7853}} \\
\textbf{\small SUDF$_{HF-Slic}$}  & \textcolor{red}{\textbf{0.8602}}   & \textcolor{red}{\textbf{0.5829}} & \textcolor{red}{\textbf{0.4671}}      & \textcolor{red}{\textbf{0.5654}} & \textcolor{red}{\textbf{0.4668}} & \textcolor{red}{\textbf{0.5436}}  & \textcolor{red}{\textbf{0.4834}} & \textcolor{green}{\textbf{2.1200}} & \textcolor{red}{\textbf{1.7241}} \\

\hline

\end{tabular}

\begin{tabular}{l}
\small *SGC \cite{8} The codes were not available for \cite{8} so the implementation was done by the authors in Matlab based on the paper \cite{8} \\ 
\small **HS-MR \cite{24} saliency detection is originally for color images; however, published codes by the authors can be used for\\ \small hyperspectral data for MR and superpixel methods in the code.
\end{tabular}
\label{tab:Metrics_comp}
\end{table*}

To achieve salient object detection goal in Fig. \ref{fig:model}, we propose to use an unsupervised backpropagation semantic segmentation method \cite{25} to learn high-level visual features that will be used in the manifold raking algorithm \cite{24} for saliency computation. Given $k$ channels hyperspectral imagery $I = \{ {H^{k}} \} _{n=1}^{k}$ as input to our model, first, all the pixel values are normalized to [0,1]. Then, we adopt a CNN model to extract p-dimension feature maps $\{ x_{n} \}$ from the Batch-Normalization (BN) output of the last Deconvolution layer of model getting hyperspectral imagery $I$ as input. The detail configuration of the CNN model is shown in Table \ref{tab:CNN_model}. Note that the spatial resolution of output feature map and input hyperspectral image $I$ are identical. After normalize the learned response maps via batch normalization as in \cite{25}, we obtain cluster label $\{ c_{n} \}$ by using argmax classification to the feature maps to classify each pixel by choosing the dimension that has the maximum value as  $\{ y_{n} \}$ \cite{25}. Then, we apply the refinement process on $\{ c_{n} \}$ based on the superpixels obtained by the high-level features of the CNN (see Fig. \ref{fig:model}) in contrary to \cite{25} using superpixels based on the input data (e.g. hyperspectral image). Refinement process is achieved by assigning all pixels same cluster label based on the highest frequency of label in the superpixel area \cite{25}.

\begin{figure}[h]
\begin{minipage}[b]{1.0\linewidth}
  \centering
  \centerline{\includegraphics[width=7.5cm]{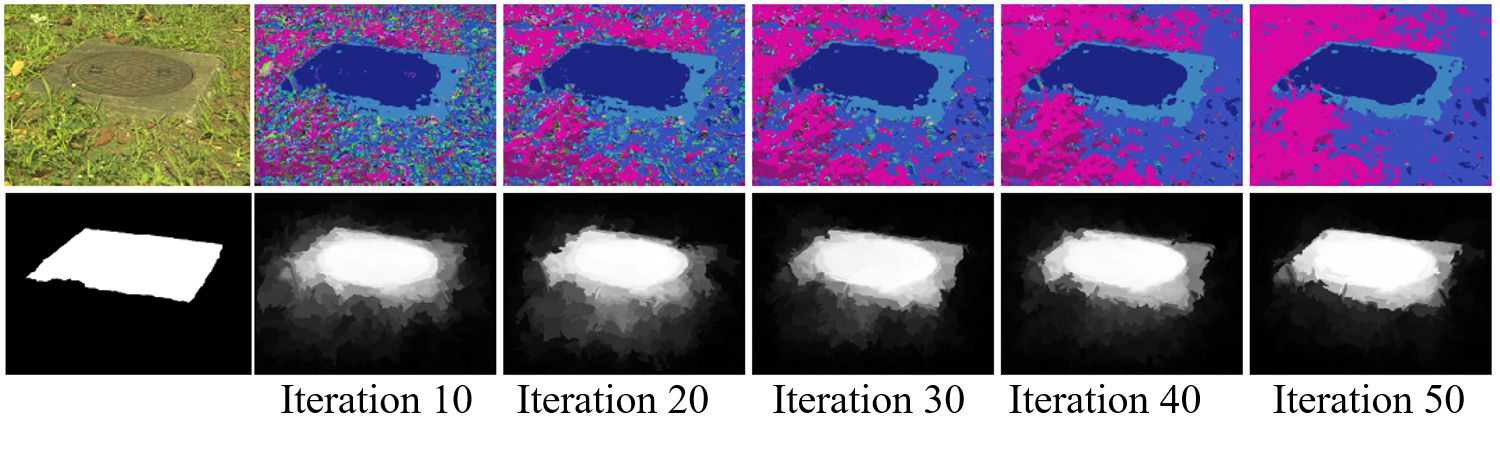}}
\end{minipage}
\caption{(top) sample image and segmentation results at different iterations during self-supervision,  (bottom) ground-truth image and saliency maps from different iterations.}
\label{fig:iteration_results}
\end{figure}

Similar with the supervised learning, we use the softmax cross entropy loss between the network responses $\{ y_{n} \}$ and the refined cluster labels $\{ c_{n} \}$ at iteration \emph{n} \cite{25}. Using this error with back-propagation, the parameters of convolutional and deconvolution filters are updated by utilizing gradient-descent with momentum \cite{25}. As in \cite{25}, Glorot and Bengio method \cite{29} is employed for the parameter initialization, which uses uniform distribution normalized according to the input and output layer size. While self-supervising the network for unsupervised segmentation task,  at each iteration,  $\{ x_{n} \}$ is used to obtain saliency map by employing MR \cite{24} with multi-channel. For the model, we use two main termination conditions as:
\begin{eqnarray}
\L_{i+1}-\L_{i} \leq \varepsilon_{1} \\
S_{i+1}-S_{i} \leq \varepsilon_{2}
\end{eqnarray}
where $\L_{i+1}$ and $\L_{i}$ denote the cross-entroppy losses of step $(i+1)$ and $(i)$, $S_{i+1}$ and $S_{i}$ denote the predicted saliency maps of step $(i+1)$ and $(i)$, $\varepsilon_{1}$ and $\varepsilon_{2}$ are defined small non-zero constants to terminate the training process. Also, when the training step $N$ achieve maximum value $\kappa=200$, it will stop the process. In Fig. \ref{fig:iteration_results}, unsupervised segmentation outputs and computed saliency maps are shown for different iterations of self-supervised learning. It can be seen that saliency map results are converging even though clustering through self-supervision is not giving optimal segmentation result yet.

\section{Experimental Results}
\label{sec:experiments}

\textbf{Metrics:} We made extensive evaluation (see Table 2) based on various performance metrics \cite{22,26,27,28} such as Area Under Curve (AUC$_{Borji}$), Cross Correlation (CC), Normalized Scanpath Saliency (NSS), Kullback-Leibler divergence (KLdiv), Precision, Recall, F-measure (F$_{\beta}$, maxF$_{\beta}$, aveF$_{\beta}$) with Precision-Recall or Precision-Recall Curves. 

\textbf{Dataset:} We made evaluation of the model on hyperspectral salient object detection (HS-SOD) dataset \cite{23} consisting of 60 hyperspectral images with their respective binary ground-truth images referring to salient objects. The dataset details can be seen in \cite{23}, and is available on \cite{29}. For each image, spatial resolution is 768x1024 pixels, and there are 81 spectral channels covering the wavelengths between 380-780nm (visible spectrum) with 5nm intervals \cite{23}. 

\begin{figure*}[!ht]

\begin{minipage}[b]{1.0\linewidth}
  \centering
  \centerline{\includegraphics[width=15cm]{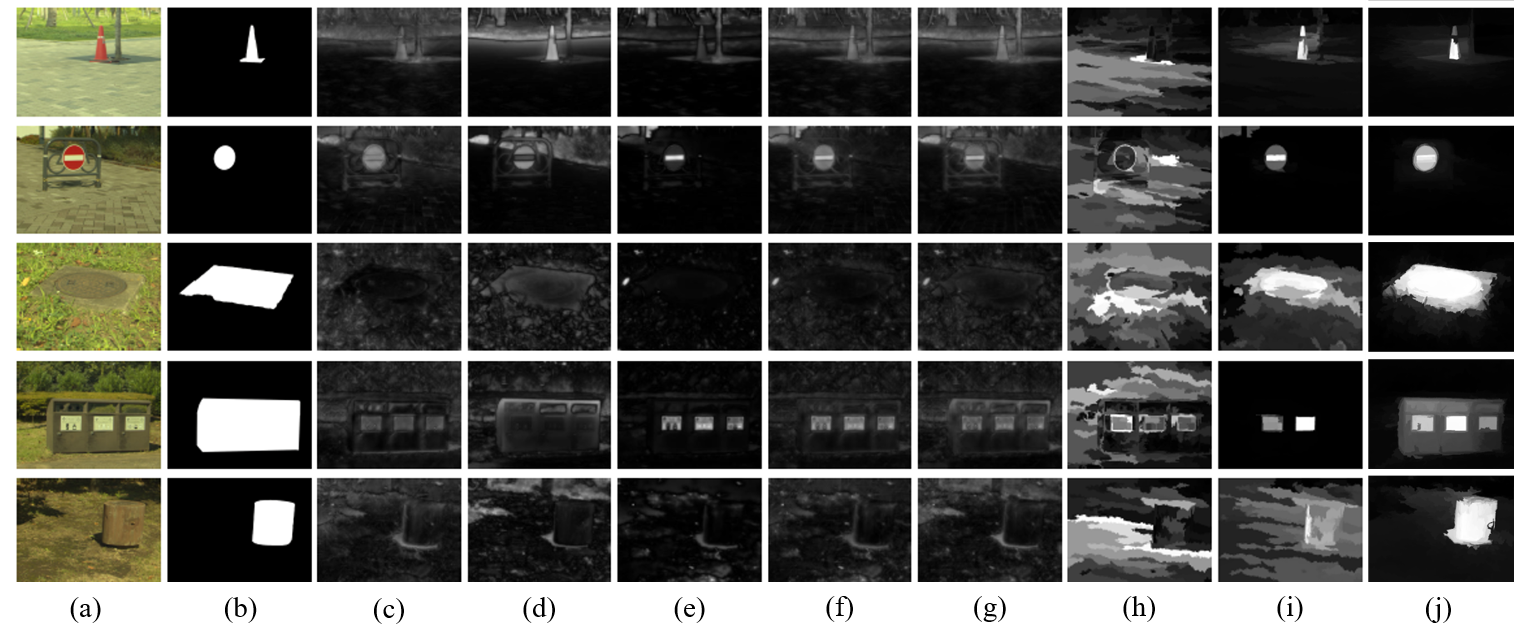}}
\end{minipage}

\caption{(a) Sample scenes of the the hyperspectral data rendered in sRGB with its respective (b) ground-truth salient objects, and saliency map results of the compard models: (c) Itti \emph{et al.}  \cite{13,3}, (d) SAD  \cite{3}, (e) SED  \cite{3}, (f) SED-OCM-GS  \cite{3}, (g) SED-OCM-SAD  \cite{3}, (h) SGC  \cite{4}, (i) HS-MR  \cite{24}, (j) \textbf{Proposed SUDF$_{HF-Slic}$}}
\label{fig:sal_results_models}
\end{figure*}

\textbf{Evaluation:} We selected \cite{3} and \cite{4} for comparison as being the hyperspectral salient object detection models for natural scenes. In work \cite{3}, various approaches were tested on hyperspectral data so we also apply the approaches tested in \cite{3} on HS-SOD dataset \cite{23} for comparison. i) spectral  distances  between  each  spatial region  for  saliency  computation  by  using  spectral  Euclidean distance  (SED)  and  spectral  Angle  distances  (SAD) \cite{3,23}, ii) color  opponency  method  in  \cite{13,3}  is  replaced  by spectral  grouping  rather  than  Red-Green  and  Blue-Yellow differences, in which Euclidean  distance  between  spectral group (GS)  vectors  by dividing spectral  bands into  four  groups  (G1,G2,G3,G4) \cite{3,23}. iii) In \cite{23}, spectral distance based saliency also combined with orientation based saliency with combinations such as SED-OCM-GS  and  SED-OCM-SAD. iv) saliency  maps  from  Itti  et  al.  \cite{13}  were also  provided for hyperspectral saliency comparison in \cite{3,23} as a  baseline  model. As a more recent work, we also tested saliency from spectral gradient  contrast  (SGC)  proposed  by  \cite{4}.  In \cite{4}, local region contrast is computed from the superpixels obtained by spatial  and  spectral  gradients, which is used to calculate spectral gradient  contrast  (SGC)  for saliency detection \cite{4, 23}. 

In addition, saliency detection by graph-based manifold ranking (MR) is also applied on hyperspectral dataset, referred as HS-MR \cite{24} to compare with the proposed model \textbf{SUDF$_{HF-Slic}$} (SUDF: Saliency from Unsupervised Deep Features ), which uses higher-level features for both MR based saliency and cluster refinement compared with the original approaches in \cite{24} and \cite{25}. In addition, to demonstrate the performance improvement on saliency detection when the cluster refinement is done on high-level fetures, we also implemented and compared saliency computation when cluster refinement is done based on input hyperspectral image that is referred as \textbf{SUDF$_{HS-Slic}$}.

As it can be seen in Table \ref{tab:Metrics_comp}, proposed \textbf{SUDF$_{HF-Slic}$} performs better than other approaches in all metrics except being second best on NSS. However, although the performance difference is very close with \textbf{SUDF$_{HF-Slic}$}, best performing model in NSS metric is also variation of the proposed approach, \textbf{SUDF$_{HS-Slic}$} which applies cluster refinement based on the superpixels obtaind from hyperspectral image directly as in the original work \cite{25}. In addition, proposed \textbf{SUDF$_{HF-Slic}$} demonstrated that using higher-level features even learned from self-supervision is more beneficial to saliency computation using manifold-ranking since proposed \textbf{SUDF$_{HF-Slic}$} outperformed HS-MR \cite{24} manifold-ranking using low-level features in all evaluation metrics. In Fig. \ref{fig:sal_results_models}, some sample scenes for hyperspectral data are given rendered in sRGB for visualization with their respective grount-truth images for salient objects, and saliency maps results of various approaches on these scenes are also given to demonstrate the performance of proposed model \textbf{SUDF$_{HF-Slic}$} with respect to other models. It can be seen from the saliency maps that our model performs better qualitatively too.

\section{Conclusion}
\label{sec:conclusion}

In this work, we demonstrated hyperspectral salient object detection approach based on self-supervised deep features in a multi-task model. Paramater update of the CNN model is done based on cross-entropy loss of clustering performance, and saliency is computed by the learned features of the unsupervised segmentation task, in which saliency convergence is the termination criteria for the self-supervised learning procedure. Evaluation on the HS-SOD dataset \cite{23} demonstrates promising results for salient object detection with the proposed approach.  As a future work, we would like to investigate how to improve representation of hyperspectral image during self-supervision process (e.g. adding sparsity loss, orthogonality constraint, decoder based image generation loss, etc. ) to improve the accuracy for salient object detection and also to increase the convergence on clustering and saliency map results. Moreover, we would like to investigate other options for saliency computation compared to MR model since it assumes boundary prior for background regions.




\end{document}